# Loopy Belief Propagation as a Basis for Communication in Sensor Networks

Christopher Crick and Avi Pfeffer
Harvard University
{crick@fas,avi@eecs}.harvard.edu

## Abstract

Sensor networks are an exciting new kind of computer system. Consisting of a large number of tiny, cheap computational devices physically distributed in an environment, they gather and process data about the environment in real time. One of the central questions in sensor networks is what to do with the data, i.e. how to reason with it and how to communicate it. This paper argues that the lessons of the UAI community, in particular that one should produce and communicate beliefs rather than raw sensor values, are highly relevant to sensor networks. We contend that loopy belief propagation is particularly well suited to communicating beliefs in sensor networks, due to its compact implementation and distributed nature. We investigate the ability of loopy belief propagation to function under the stressful conditions likely to prevail in sensor networks. Our experiments show that it performs well and degrades gracefully. It converges to appropriate beliefs even in highly asynchronous settings where some nodes communicate far less frequently than others; it continues to function if some nodes fail to participate in the propagation process; and it can track changes in the environment that occur while beliefs are propagating. As a result, we believe that sensor networks present an important application opportunity for UAI.

## 1 Introduction

Sensor networks are an exciting new kind of computer system. They consist of a large number of tiny, cheap computational devices distributed in an environment. The devices gather data from the environment in real time. Some data processing occurs in real time within the network itself; other data is shipped to a server for offline processing. In some cases the devices react online to the state of the environment.

One of the central questions in sensor networks is what to do with the data. When the data is to be processed online within the network, what form should the information take, how should it be computed, and how should it be communicated? When nodes need to react to the information online, how can we ensure that each node receives the information it needs? In addition, how should the overall flow of information be organized? All this needs to be accomplished at minimal cost and in a distributed fashion.

Consider, for example, the task of monitoring a building for outbreak of fire. A set of temperature sensors will be deployed throughout a building. Accurately detecting fire requires combining information from multiple sensors. For example, if a fire breaks out midway between two sensors, combining slightly elevated temperature readings at each of the sensors can provide a much quicker response than waiting until a single sensor has a very high reading. In addition, sensors that are physically deployed for a long time in an environment are subject to multiple kinds of failure. They may provide noisy readings, or they may break down completely. Combining information from multiple sensors can overcome these types of failure. In this application, we would like an immediate online response to occur as soon as a fire is strongly believed to be happening. Ideally, the sensor information would be combined online, to produce a quick and accurate response. How is this to be done?

This paper argues that the UAI community can provide good answers to these questions. In many applications, like fire monitoring, the key task is to form beliefs about the state of the system based on the collected sensor readings. Since the environments in which sensor networks are deployed typically have a great deal of uncertainty, this is a core UAI task.

In particular, we argue that loopy belief propagation (LBP) is an ideal computational and communication framework for sensor networks. LBP has emerged as one of the leading methods for approximate inference in graphical models. It has properties that make it naturally suited for the sensor network domain. It can easily be implemented as a distributed algorithm, and the processing performed at each node is very simple and can be implemented cheaply in a tiny device.

The sensor network application presents many challenges to the LBP framework that have not been encountered in previous applications. We investigate experimentally whether LBP is able to withstand some of these challenges. First, algorithms for sensor networks should be asynchronous. Attempting to enforce synchrony and a particular order of processing would be costly, and could lead to a loss of robustness if one step of processing fails. The first step of our investigation is to confirm that LBP does not rely on a synchronized order of message passing, but works just as well in an asynchronous environment in which each node communicates intermittently. Second, sensor networks often consist of devices of vastly different size and computational capability, and in addition the devices deployed in a physical system may be at very different levels of functionality. As a result, we would expect that in a deployed system, there will be nodes that compute and communicate far more frequently than others. We show that LBP continues to perform well even in highly asynchronous systems with vastly different communication rates. Third, nodes in a sensor network are subject to failure. Good sensor networks are designed with redundancy to allow for such failure. We show that LBP can exploit such redundancy to perform well even as nodes fail, and that it enjoys a graceful degradation property. Fourth, in LBP, beliefs gradually converge to the correct beliefs after a change in sensor readings. In a dynamic setting, it is possible that the environment might change again before the beliefs have had a chance to converge. One might suspect that this would lead to an unstable system, where the beliefs never track the truth. We show that this is not the case, and that LBP continues to perform well even when we expect many environmental changes to occur in the time it takes beliefs to converge.

As a result of our experiments, we assert that LBP is a strong candidate to be a basis for computation and communication in sensor networks. It is semantically well founded, computing correct beliefs from sensor readings, relative to a probabilistic model. It is simple enough to be deployed in a wide variety of domains. Most important, it enjoys a number of properties that allow it to cope with the stresses of deployment in a dynamic system subject to various kinds of failure.

## 2 Sensor Networks

The push toward sensor networks has been driven by advances in hardware [9]. Silicon devices can be made smaller and cheaper than ever before. As a result, one can now envision systems that rely on hundreds, thousands or even more of these devices. These systems require a totally new approach to large-scale computing. On the one hand, they are much more tightly integrated with the environment than previous systems. Instead of relying on a small number of interfaces, every tiny piece of the system is embedded with and in contact with the environment. On the other hand, they rely fundamentally on computation being done by a large number of distributed devices that individually have limited capability. The level of devices varies from low-power, low-functionality devices to those that use a small operating system such as TinyOS [4] to achieve a reasonable level of computational capability. Algorithms for these devices must be implemented very cheaply, perhaps directly in hardware.

Sensor networks have a wide variety of potential applications. One type of application is simple data collection for the purpose of offline study. For example, one proposed network will collect data for studying Sudden Infant Death Syndrome (SIDS) by placing sensors in diapers. In such applications, where the main purpose is collecting data for offline data mining, there is no need for data processing to form beliefs online. In other applications, however, the purpose is real-time response. In addition to the fire-monitoring application, one can consider a disaster recovery application in which the goal is to help hospital services cope with a large-scale disaster. More mundanely, one might imagine a widespread, fine-grained inventory control system. In such applications, the system would be greatly enhanced by the ability to form beliefs online.

Another application of sensor networks has been in the field of robotics. Decentralized sensor systems have been used in automated navigation and tracking in a variety of environments [2]. Such decentralized data fusion increases the scalability, survivability and modularity of a robot by eliminating critical points of failure. Current systems rely on a distributed Kalman filter algorithm for computing the local information at each node. However, such an approach requires that the network be tree-connected, which rules out many useful applications.

The term "sensor networks" is a misnomer, since there are many other kinds of devices in the systems. In addition to sensors, the devices may contain actuators that exert control on the environment. One proposed design for sensor networks, the Hourglass architecture [12], envisions three additional kinds of sensors: data nodes are equipped with a small amount of stable storage and are responsible for collecting and storing data from sensor nodes; communication nodes are responsible communicating with the outside world, which is a relatively expensive operation; and processing nodes perform some computation on the data within the network itself. The key point is that labor is divided between different kinds of devices. The sensor nodes themselves are not expected to do a lot of processing. In this paper, we focus on the sensor nodes that directly collect information and the processing nodes that use the information to compute beliefs online. We would expect each processing node to examine a set of sensor nodes, and communicate with a small number of other processing nodes.

If construed in the widest sense, we can also consider the term sensor network to include virtual networks of sensors distributed across the internet. Such a network could be useful for internet security. For example, one of the major current types of security problems are distributed denial-of-service (DDoS) attacks, where an attacker floods a site with so much traffic that it effectively cuts the site off the internet. One of the best approaches developed thus far for combating DDoS is pushback [5]. When a router notices traffic passing through it above a certain threshold, it holds up the traffic, and also notifies the upstream source. This is only a local response. Using a sensor network approach, one could potentially develop a global response. Consider a system in which a small percentage of the routers in the internet try to monitor the level of traffic to target sites, and periodically send messages to each other. The number of such messages would be very small relative to the overall amount of internet traffic, so would require very little overhead. However, using such messages, each of the routers could form beliefs about the existence of a DDoS attack against a site. This would have multiple advantages. First of all, as soon as the DDoS attack is detected, all the participating routers could engage in pushback, greatly increasing the effect of the response. Equally important, the DDoS attack could potentially be detected much sooner. Instead of having to wait for a single node to see traffic above a high threshold, the attack could be detected as soon as a large number of nodes see traffic that is only somewhat above threshold. The ideas of this paper, about using loopy belief propagation as a basis for forming and communicating beliefs, hold equally well for such a virtual sensor network as for a physical network.

## 3 Modeling a Sensor Network as a Graphical Model

There is a great deal of uncertainty in any sensor network system. For one thing, the underlying domain exhibits uncertainty; without it, we would not need to deploy sensor networks throughout the system. The sensors typically give us only partial information about the state of the system; otherwise, we would not need to compute beliefs, we would know the answers simply by looking at the sensors. Furthermore, the sensors are noisy, they might be biased, and they might be broken. In short, reasoning under uncertainty to form coherent beliefs is a major task in sensor networks.

Each sensor individually provides a reading for a particular state variable at a particular point. That reading may depend not only on the system state but also on the sensor properties. Any interaction between sensors is assumed to be the result of high-level variables. It is the job of the processing nodes to form beliefs about these high-level variables from the sensor readings, taking into account the possibility of sensor error. In a complex, widely distributed system, there will be many interacting high-level variables, and their interactions might form a complex, loopy network. Each processing node will be responsible for a set of sensors and a small number of high-level variables. We can draw a network of processing nodes, in which two nodes are neighbors if their high-level variables interact. We now make a key assumption: that the the joint probability distribution over the states of all processing nodes can be decomposed into the product of *pairwise interactions* between adjacent processing nodes. This might be an approximation, but we believe that for physical systems that operate through local interactions it will tend to be a good one.

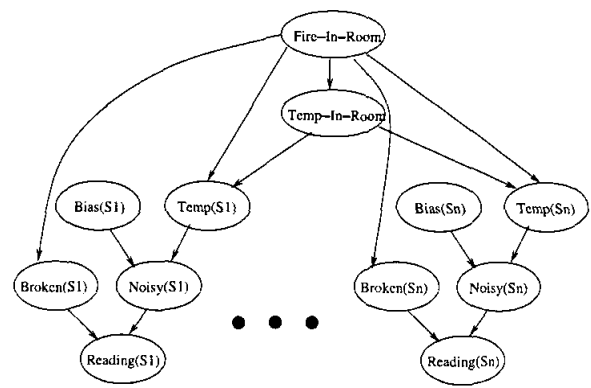

Figure 1: Local BN for processing node

For example, suppose we have a temperature sensor $S$ at some location. We model READING($S$) as depending on TEMP($S$). In addition, $S$ will have a certain bias BIAS($S$) which is added to the temperature to produce NOISY($S$). However, there is also a BROKEN($S$) variable; if $S$ is completely broken, the reading will be random. Each processing node will be responsible for a set of sensors. The processing node will have high-level variables TEMP-IN-ROOM representing the ambient temperature and FIRE-IN-ROOM, a Boolean representing whether or not there is a fire in the room. FIRE-IN-ROOM naturally influences TEMP-IN-ROOM, and TEMP-IN-ROOM influences TEMP($S$) at each of the sensor locations. In addition, FIRE-IN-ROOM influences TEMP($S$) because the temperature at a point is likely to deviate more from the ambient temperature if there is a fire. FIRE-IN-ROOM also makes it more likely that a sensor is broken. A schematic of the Bayesian network for a single processing node is shown in Figure 1.

In addition, the temperature in adjacent rooms is highly correlated, as is the existence of fire. Therefore the TEMP-IN-ROOM and FIRE-IN-ROOM variables associated with one processing node are connected to those of neighboring processing nodes. Since the relationship between adjacent processing nodes is symmetric, a Markov network is more appropriate than a Bayesian network for capturing the connectivity at this level. The relationship between two adjacent processing nodes is modeled with a compatibility function, which will be higher the closer the values of TEMP-IN-ROOM and FIRE-IN-ROOM between the two nodes. The graph of processing nodes, corresponding to the adjacency graph of locations, will be quite loopy.

The inference task in a processing node is to compute the distribution over high-level variables given sensor readings. No information need be passed back to the sensors. The sensors do not need to be told whether they are broken; that possibility is taken care of at the processing node. Since the network is used for a specific query, a technique such as Query DAGs [1] can be used to produce a computational framework in which the local beliefs can be computed very quickly. Query DAGs were designed for implementation in software or hardware for on-line, real-world applications, and so are ideal for computing local beliefs within a single processing node. However, they cannot be used for the overall process of computing beliefs in sensor networks, since they rely on exact inference algorithms. Since the inter-processing-node network can be quite loopy, exact algorithms are infeasible.

## 4 Loopy Belief Propagation

We therefore need to use an approximate inference algorithm. Furthermore, we need one that can easily be implemented in a distributed form, and that can be implemented efficiently in software or hardware. Loopy belief propagation (LBP) fits both of these criteria.

LBP is an extension of the belief propagation framework developed by Pearl for the polytree algorithm [10]. Pearl in fact emphasized the distributed potential of the algorithm as one of its attractive properties. While the algorithm produces exact beliefs in singly-connected networks, it does not do so in networks with loops. Pearl discussed the idea of running belief propagation in loopy networks, but expressed concern that the beliefs would not converge.

In the coding community, the hugely successful Turbo coding scheme was developed, and it was shown that its decoding algorithm is equivalent to running belief propagation in a loopy network [7]. As a result of this success, there has been a resurgence of interest in the use of LBP as a general approximate inference algorithm for Bayesian networks. Empirical studies [8, 11] have shown that LBP is a highly competitive approximate inference algorithm. It works very quickly, and generally producing accurate approximations to the correct beliefs. While the algorithm does not always converge, cases of non-convergence are relatively rare and can easily be detected. Meanwhile, recent work [6] has laid the theoretical foundation for understanding LBP as well as pointing to generalizations. Due to its ease of implementation and strong empirical performance, LBP is emerging as a leading algorithm for approximate inference.

We follow [6] in our presentation of LBP. The nodes in the algorithm are the processing nodes. The value $x_i$ of node $i$ is the state of the high-level variables of processing node $i$. Let $y_i$ denote the sensor readings at $i$, and $\mathbf{y}$ the complete set of sensor readings. The complete joint distribution over the state of the system, given the sensor readings, can be expressed as

$$P(x_1, \ldots, x_N \mid \mathbf{y}) = \frac{1}{Z} \prod_{ij} \psi_{ij}(x_i, x_j) \prod_i \phi_i(x_i \mid y_i)$$

where $Z$ is a normalization constant, the first product is taken over adjacent nodes, $\psi_{ij}$ is the compatibility function between nodes $i$ and $j$, while $\phi_i$ represents the effect of the local sensors on the belief in node $i$, as computed by the BN in node $i$. In LBP, each node $i$ sends a message $m_{ij}$ to each of its neighbors $j$, and updates its beliefs $b_i$ based on the messages it receives from its neighbors. The update rules are:

$$m_{ij}(x_j) \leftarrow \alpha \sum_{x_i} \psi_{ij}(x_i, x_j) \phi_i(x_i \mid y_i) \prod_{k \in N(i) \setminus j} m_{ki}(x_i)$$

$$b_i(x_i) \leftarrow \alpha \phi_i(x_i \mid y_i) \prod_{k \in N(i)} m_{ki}(x_i)$$

where $\alpha$ is a normalization constant, $N(i)$ denotes the neighbors of $i$, and $N(i) \setminus j$ denotes the neighbors of $i$ except for $j$. The belief at a node takes into account the local evidence at the node, and the messages sent to it by all its neighbors. The message a node $i$ sends to its neighbor $j$ tells $j$ which values it thinks are likely for $x_j$, based on what $i$ thinks is likely for $x_i$ as a result of its local evidence and messages from other neighbors, and the compatibility between $x_j$ and $x_i$.

## 5 Asynchronous Behavior

Since LBP is defined in terms of local update rules, it can easily be implemented in a distributed fashion. Furthermore, the algorithm requires only a simple set of multiplications and additions which can easily be implemented on a tiny device. In addition, the algorithm as formulated does not rely on any coordination of the messages. Each node can update its own beliefs and the messages it sends to its neighbors at any time, using the most recently sent messages from its neighbors. In practice, however, implementations of LBP on sequential computers have been synchronous. The simplest way to run LBP on a sequential machine is for nodes to take turns updating and sending messages.

While there was no reason in principle to believe that LBP would not work equally well in an asynchronous environment, the possibility existed that the surprising convergence of LBP relied on an organized propagation schedule. If that was the case, any attempt to apply LBP to sensor networks would be doomed to failure. We therefore began our experimental investigations by determining whether the convergence properties found in previous experiments held up for an asynchronous implementation.

We performed experiments using two real-world Bayesian networks, ALARM and HAILFINDER [3], and a synthetic sensor network called FIRESENSOR based on the fire monitoring model described in Section 3. The two real-world nets are fairly small – 37 and 56 nodes, respectively, while FIRESENSOR consists of 680 nodes modelling one hundred identical sensor clusters connected in a 10x10 lattice. In each experiment, between 0 and 20% of the nodes were randomly assigned an observed value. We found that in all three networks, LBP continues to perform well under asynchronous conditions. In particular, asynchronous LBP converged whenever the synchronous version did, and to the same beliefs.

Next, we investigated whether LBP was robust to wide variations in the rate at which nodes communicated. A typical sensor network will consist of devices of very different levels of capability, that compute and communicate at very different rates. Furthermore, devices in a deployed system will tend to adjust their computational performance to circumstances. For example, as a device loses power it will tend to communicate less frequently. In addition, it may be more difficult for one device to communicate to another for environmental reasons. For instance, if there is a lot of interference the signal from a device may only be picked up intermittently. We can model this situation as one in which the device communicates less frequently. For a variety of reasons, then, we can expect communication in a sensor network to happen at very different rates. LBP relies on all the nodes updating their beliefs and communicating messages to their neighbors. Is it able to cope with different rates of communication?

To answer this question, we ran experiments in which half the nodes in the network are much more likely to propagate than the other half. Each node used a exponential random process to determine when to pass messages to its neighbors. In a typical experiment, half the nodes were 10 times more likely to propagate than the others. Other ratios were also tested with similar results. We found that in all cases, asynchronous belief propagation with different propagation rates converged to the correct beliefs whenever ordinary LBP did.

In addition, the asynchronous cases perform much better than expected in terms of the number of propa-gations required for beliefs to converge. For instance, it might take $cn$ propagations for the network to converge in the nondistributed algorithm, where the constant $c$ depends on the topological characteristics of the network – for instance, how many times messages must pass around a loop before its constituent nodes' beliefs converge. We might expect, based on this, that all nodes need to propagate about $c$ times to achieve convergence. If this is the case, basic probability tells us that for an asynchronous network with uniform propagation times, it would take an expected $cn \ln n$ propagations to converge. Figure 2 shows a comparison of the number of propagation times required until convergence for synchronous LBP, asynchronous LBP with uniform propagation times, and asynchronous LBP with different propagation times. For each test network and each algorithm, the graph shows the minimum, maximum and median number of propagations required until convergence. In general, we see that the uniform asynchronous LBP uses only $\frac{2}{3}$ as many propagations as we might expect. This indicates that some nodes can underreport and the correct beliefs are still reached.

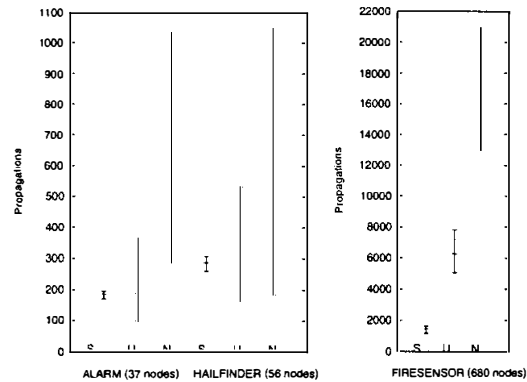

Figure 2: Convergence performance in synchronous, uniform asynchronous, and non-uniform asynchronous networks

As for LBP with different propagation times, while the total number of propagations required is significantly more than for synchronous LBP, it is far less than one would expect if the slow-propagating nodes had to propagate $c$ times in order for beliefs to converge. In a network in which half the nodes propagate 10 times more often than others, we would expect the total number of propagations required for the slow-propagating nodes to propagate $c$ times to be about 5 times as high as for a uniform network. In fact we find performance to be much better. For example, in the FIRESENSOR network the median number of propagations in the non-uniform case is 16,000, compared to 6,000 for the uniform case. It must be that by continuing to propagate, the fast-propagating nodes are "working overtime" and making up for some of the lack of propagation of the slow-propagating nodes. This is a nice property: as some of the nodes in the network slow down, they only partially slow down the network as a whole. We also discovered that the speed of convergence varies widely based on which nodes propagate often. Nodes that are centrally located and have large impact on the network also have a profound effect on inference speed. For example, if the 10 most highly-connected nodes (meaning the ones with the most parents and children) of the ALARM network are set to propagate three times as often as the rest of the nodes in the network, this distributed, asynchronous process converges in just about the same number of steps as the synchronous LBP algorithm. These results suggest that, when building sensor networks, identifying central nodes and applying resources to increase their speed would have a disproportionate positive effect on overall system performance.

## 6 Robustness to Failure

The fact that networks continue to converge even when certain nodes participate markedly less often than others leads to the natural question of how such networks perform when some nodes fail to participate at all. In a distributed system of simple devices, some will fail, and one would hope that such failures are not fatal. One can distinguish between different kinds of failures, corresponding to different kinds of nodes. Failure of sensor nodes are handled naturally in the probabilistic model by the BROKEN variables. This is a familiar kind of failure in probabilistic reasoning. Less familiar, however, is failure of propagation nodes, which results in nodes ceasing to participate in the belief propagation process. Not only do they not form beliefs about their own state variables, they fail to send messages to other nodes. This could potentially ruin the LBP process. In fact, however, our experiments show that LBP continues to function in the face of "dead" nodes and degrades gracefully as their numbers increase.

Network topology has a profound effect on the performance of loopy propagation under degraded conditions. Redundancy is crucial. Without it, a single point of failure will cause the whole system to break down. In the limiting case, a node that bisects a network plays a crucial role in establishing accurate system-wide beliefs. If such a node ceases to function, then evidence on one side of the network cannot affect beliefs on the other side, and neither subnetwork can arrive at accurate beliefs. However, as long as at least one alternate path for information flow exists, inference is remarkably resilient.

We performed two sets of experiments with our synthetic sensor network to study the effects of node degradation. In the first, we randomly selected sensors, from 2 to 20 out of the 100 in the total network. We rendered them inoperable, then compared the beliefs of working nodes to their counterparts in a fully functional converged network. We randomly assigned observations to 10% of the nodes, choosing from among the working ones. We identified the number of nodes in the degraded network that differed from the values produced by the fully operational one, and determined the magnitude of the belief difference.

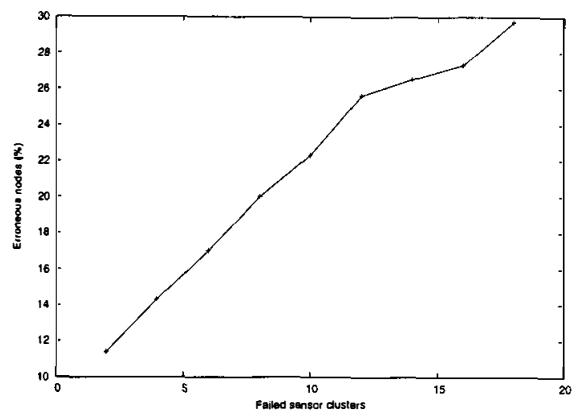

Figure 3: Network degradation resulting from random sensor failures

Figure 3 shows the performance degradation as the number of failed nodes increases. With two sensors nonfunctional, only nodes directly connected to the problem sensors show any errors at all, about 12% of the total network. As more nodes go offline, the number of affected nodes increases, but even with a fifth of the nodes dead, nearly two thirds of the network remains untouched by the problems.

Most of the nodes in a degraded sensor network remain completely reliable. The affected nodes, on the other hand, can be somewhat off the mark, but their beliefs still tend in the right direction. Somewhat unexpectedly (but presumably coincidentally), the average absolute belief error among the affected nodes remains almost perfectly constant as the number of dead sensors increases – right around 13%. The largest errors are found in nodes close to dead sensors and far from any observed evidence; the smallest are in those nodes strongly influenced by observations.

It is not surprising that nodes neighboring broken ones should have somewhat degraded performance, since they lose crucial information in forming their beliefs. It is also not surprising that nodes far away from observations should be more seriously affected than those close to them, since such nodes depend more heavily on their broken neighbor. What is perhaps more surprising is that the degradation does not spread in a significant way to neighboring nodes beyond the affected nodes. The beliefs at the erroneous nodes are accurate enough that their neighbors are able to obtain most of the information they need. Thus loopy belief propagation has a highly appealing graceful degradation property. Not only is the degradation local as nodes in the network fail, but it dissipates very quickly as one moves away from the failed nodes.

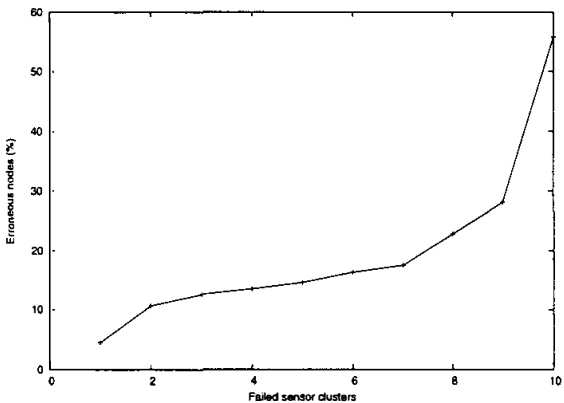

Figure 4: Network degradation resulting from path-blocking failures

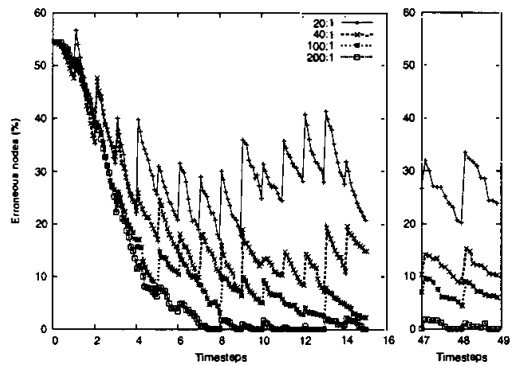

Figure 5: Convergence performance at propagation speeds relative to environmental change

We ran a similar experiment on a variation of FIRESENSOR in which the processing nodes were connected in the following pattern:

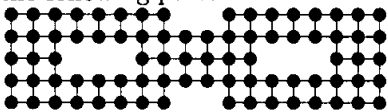

The network was designed to look like a plausible floor-plan for a building, which might not have as much redundancy as the complete lattice. Results were similar in nature to FIRESENSOR, and only slightly worse, due to the increased probability that knocking out a few nodes will block all paths from one side of the network to the other.

We designed our second set of experiments to explore the importance of redundant propagation paths in maintaining accurate beliefs, and the results are summarized in Figure 4. Instead of choosing nodes randomly, we removed one sensor at a time from the fifth row of the 10x10 network, so that the number of communication paths between the bottom and the top of the network steadily decreased. Just as in the last experiment, we measured the total number of nodes with incorrect beliefs at convergence. Until we killed the eighth node, leaving only two paths, the system's performance did not differ significantly from the random case. Even then, the number of affected nodes was only marginally worse, by about 10%. Of course, once all ten nodes in a row fail, error becomes extreme – only nodes whose beliefs are entirely determined by local observations reach correct beliefs. Thus individual messages between nodes in loopy propagation seem to encode an enormous amount of information about the state of large swaths of network – as long as a path exists, beliefs will flow.

## 7 Dynamic Behavior

Sensor networks are not static entities; their whole function is to change and adapt to fluctuations in the environment they are measuring. Although asynchronous networks propagate beliefs quickly and efficiently, they cannot do so instantaneously. If the environment is changing rapidly, we may encounter situations in which it changes several times in the time necessary for beliefs to converge. As a result, the whole belief propagation process could potentially become unstable. We performed experiments testing LBPs ability to adapt in a dynamically changing environment. Happily, we found that even when nodes make and change observations in the midst of loopy propagation's flurry of messages, a system's overall beliefs continue to converge to accurate values.

In our experiments, we varied the rate of environmental change as a function of propagation time. We define a *time step* to be the mean propagation interval for each node, using an exponential random propagation model with a uniform mean across all of the nodes. At each time step, every node has a small chance of making a fresh observation. We simulated runs of the system, and measured performance as follows. Every $\frac{1}{10}$ of a time step, we determined the number of nodes whose beliefs differ from what they would be if the network had enough time to converge fully with a given set of observations. We compare the beliefs to those that would be obtained with an "instantaneous" LBP, rather than the true correct beliefs. This provides a measure of the error in the system due to slow convergence, as opposed to error due to the LBP approximation. To obtain an overall measure of performance we averaged this error over different time points.

Suppose that at each time step, each node makes a fresh observation with probability of $p$. Then there will be $np$ environmental changes in each time step. For example, we found that for FIRESENSOR the network converges fully at nearly every time step if $p = 0.005$.

Since the network has 680 nodes, over 3 environmental changes happen at every time step. Since this network ordinarily converges in 9 time steps with no environmental changes, about 30 such changes happen in the convergence time of the network. Nevertheless, the network has very low error in steady state. Even when environmental changes occur 10 times more rapidly, only about 20% of the network holds incorrect beliefs at any particular time step.

Figure 5 shows the percentage of incorrect nodes in the sensor network over time at various rates of environmental change. We see that after a short burn-in period, in which the beliefs converge from their initial random settings, the percentage of incorrect nodes remains fairly stable in steady state. We conclude that LBP performs well even as the environment changes rapidly. Furthermore, it remains stable as the the speed of environmental change increases, with graceful degradation in performance.

## 8 Conclusion

We view the contributions of this paper as threefold. The first is to identify sensor networks as a fitting application area for probabilistic reasoning technology. Sensor networks are an exciting and growing field, and questions naturally arise there that have been studied by UAI researchers for years.

Secondly, we identified LBP as a particularly appropriate technology for sensor networks. It is fast and has shown to generally produce accurate results. It is naturally distributed, and can easily be implemented in an asynchronous environment. It is also very simple, and the computations required at every node can be implemented in low-level software or hardware.

Thirdly, we have run a series of experiments to test whether LBP can withstand the variety of stresses that would be placed on it in a sensor network environment. LBP came through the experiments with flying colors, in fact surpassing our expectations. It continued to perform with highly non-uniform propagation, and in addition, required far fewer propagations than expected. Not only did it continue to work in the presence of node failures, but problems remained confined locally to the failure region. Stable in the face of environmental changes, it even continues to perform when many such changes occur during the time it takes to converge. As a result of our experiments, we believe that LBP is up to the task of providing a foundation for reasoning and communication in sensor networks.

One issue not addressed in this paper is modeling the system dynamics. Our graphical model is a snapshot model of the state of the system at a particular point in time, and does not model changes in the state. While we have shown that beliefs in the static model are able to adapt to changes in the environment, we might do better by explicitly modeling and reasoning about such changes, using a representation such as a dynamic Bayesian network. Extending our framework to such models is a topic for future work.

## Acknowledgements

We would like to thank Margo Seltzer for discussions on sensor networks. This work was sponsored by ARO grant DAAD19-01-1-1610.